\documentclass[runningheads]{llncs}

\usepackage{graphicx}
\usepackage{xcolor}
\usepackage{enumitem}
\usepackage{tabularx}
\usepackage{makecell}
\usepackage{comment}
\usepackage{array}
\usepackage{caption}
\usepackage{multirow}
\usepackage{color, colortbl}
\usepackage{bm}

\begin{document}
\title{Assessing Generative Language Models in Classification Tasks: Performance and Self-Evaluation Capabilities in the Environmental and Climate Change Domain
}
%
%
\author{Francesca Grasso\thanks{Corresponding Author.}\orcidID{0000-0001-8473-9491} \and Stefano Locci\orcidID{0009-0006-9725-2045}}
\authorrunning{}
%
\institute{Department of Computer Science, University of Turin\\
Corso Svizzera 185 - 10149 Turin, Italy\\
\email{\{fr.grasso, stefano.locci\}@unito.it}}
%
\maketitle 
\begin{abstract}
This paper examines the performance of two Large Language Models (LLMs) - GPT-3.5-Turbo and Llama-2-13b - and one Small Language Model (SLM) - Gemma-2b, across three different classification tasks within the climate change (CC) and environmental domain. Employing BERT-based models as a baseline, we compare their efficacy against these transformer-based models. Additionally, we assess the models’ self-evaluation capabilities by analyzing the calibration of verbalized confidence scores in these text classification tasks. Our findings reveal that while BERT-based models generally outperform both the LLMs and SLM, the performance of the large generative models is still noteworthy. Furthermore, our calibration analysis reveals that although Gemma is well-calibrated in initial tasks, it thereafter produces inconsistent results; Llama is reasonably calibrated, and GPT consistently exhibits strong calibration. Through this research, we aim to contribute to the ongoing discussion on the utility and effectiveness of generative LMs in addressing some of the planet’s most urgent issues, highlighting their strengths and limitations in the context of ecology and CC.
\end{abstract}
\section{Introduction}

The advent of Large Language Models (LLMs) in both practical applications and academic research has marked a significant paradigm shift in the Natural Language Processing (NLP) community's focus towards these models and their potential. Recent years have witnessed a surge in studies employing, assessing, evaluating, and leveraging LLMs for various objectives \cite{Chang2023ASO}. These include traditional NLP tasks like sentiment analysis \cite{Zhang2023SentimentAI} 
and text classification \cite{Sprenkamp2023LargeLM}, among others. Despite their remarkable ability to generate contextually consistent outputs using natural language and enable a broad range of tasks 
—from straightforward proofreading to more complex challenges like generating code for algorithms— 
LLMs are not without their flaws, being prone to errors, misunderstandings, and hallucinations \cite{mittal2024towards}.
In particular, when it comes to task- or domain-specific performance, especially in classification tasks, LLMs demonstrate conflicting results as compared to other well-grounded machine learning techniques such as transformer-based models 
and Naive Bayes \cite{Yu2023OpenCO}.
If some works highlight LLMs' proficiency in achieving comparable results to the current state-of-the-art \cite{Sprenkamp2023LargeLM,Clavi2023LargeLM}, other studies show that their extensive capabilities may lead to suboptimal performance against
more specialized 
models \cite{Chen2023EvaluationOC,Sun2023TextCV}.
These observations have 
driven the NLP community towards exploring also 
small language models (SLMs), noted for their compact architecture and fewer parameters compared to larger models \cite{Yu2023OpenCO}.

When it comes to domain-specific applications, despite the extensive discourse surrounding LLMs, there remains a scarcity of exploration into their application for environmental and climate change (CC)-related texts, a domain of growing significance within the NLP field \cite{Stede2021TheCC,Hershcovich2022TowardsCA}. The urgency of addressing CC and ecological crisis through NLP techniques has begun to emerge as a crucial point of research. This includes tasks such as sentiment analysis 
\cite{MohamadSham2022ClimateCS}, and stance detection 
\cite{Upadhyaya2022AMM}. Moreover, the scope of interest should further broaden to encompass a wider array of environmental issues, from deforestation to plastic pollution, underscoring the importance of this research beyond CC to a more holistic environmental perspective \cite{Ibrohim2023SentimentAF,grasso-etal-2024-ecoverse}.

To address these issues, we aim to explore the efficacy of open and close Large and Small Language Models (L/SLMs) in three ecology-related text classification tasks (Eco-Relevance, 
Environmental Impact Analysis, and Stance Detection)\cite{grasso-etal-2024-ecoverse}. 
In particular, we utilize three representative generative models for these categories:
GPT-3.5-Turbo-0125 (closed LLM); Llama-2-13b-chat-hf (open mid-LLM), Gemma-2b-it (open SLM), and compare their performance with the baseline consisting of the results for the same tasks of three small, non-generative language models.
Moreover, this assessment extends to examining the models' self-evalution capabilities through the analysis of their prompt-elicited confidence scores to measure their calibration levels\footnote{The code is available here: \url{https://github.com/stefanolocci/LLMClassification/}}. Self-evaluation, the ability of language models to assess the validity of their own outputs, has proven to be crucial for enhancing accuracy and content quality when properly calibrated \cite{Ren2023SelfEvaluationIS}.
In summary, we pose three research questions:

RQ1: How accurately do state-of-the-art generative L/SLMs, both open-source and closed-source, identify texts related to ecology, analyze environmental impacts, and determine the stance 
on environmental issues?

RQ2: Compared to a baseline set by BERT-based classifiers, do these models demonstrate comparable accuracy in classifying texts within the domain of environment and CC?

RQ3: How well-calibrated are the verbalized confidence scores of LLMs and SLM in these environmental classification tasks?
\\Our results indicate that while LLMs generally outperform the SLM, BERT-based models still generally surpass both LLMs and SLM. Notably, the performance of large generative models, especially GPT, is significant, particularly in terms of recall. The calibration assessment revealed mixed results: GPT produced well-calibrated outputs, LLama exhibited moderate calibration, and Gemma struggled to adhere to the prompt template within multilabel settings. These findings aim to contribute to our understanding of the capabilities of L/SLMs in environmental text classification, enriching the ongoing discussion on leveraging NLP for ecological and CC research.

\section{Related Works}
Given the considerable amount of work and vertiginous growth of material on the topic of LLMs, 
we report a limited number of studies relevant to our research.
\textbf{LLMs in Classification Task}
Text Classification, a fundamental NLP task, has progressed from traditional methods such as Naïve Bayes \cite{McCallum1998ACO}, 
to advanced models like BERT \cite{Devlin2019BERTPO,liu2020roberta}.
Recent studies on LLMs in text classification present mixed outcomes when compared to traditional supervised methods. For instance, \cite{Yu2023OpenCO} explores the performance of different classes of language models in text classification, often favoring smaller models. \cite{Arco2023LeveragingLV} investigates LLMs across languages and tasks without a clear best performer, while \cite{Chen2023EvaluationOC} discusses ChatGPT's limitations in biomedical fields compared to domain-specific models. Nonetheless, advances in prompting techniques, such as Clues and Reasoning prompts \cite{Chen2023EvaluationOC}, and strategic prompts \cite{Clavi2023LargeLM}, show potential in improving LLMs' classification abilities. \cite{Sprenkamp2023LargeLM} highlights GPT-4's comparable performance in propaganda detection to the current state-of-the-art.
\\\textbf{LLMs 
Self Evaluation}
The significant hype around LLMs has encouraged numerous studies on their capabilities \cite{Chang2023ASO} and self-evaluation skills, including calibration—the alignment between a model's confidence and its prediction accuracy. Calibration is crucial for a model's reliability, especially in determining when to defer to an expert \cite{Tian2023JustAF}. Studies \cite{Tian2023JustAF,Lin2022TeachingMT} indicate LLMs' verbalized confidences are often better calibrated than conditional probabilities, yet \cite{Xiong2023CanLE,Ren2023SelfEvaluationIS} highlight their tendency towards overconfidence. Our research evaluates the accuracy and calibration quality of LLMs' verbalized confidence scores.
\\\textbf{LLMs in Environmental Domain}
Regarding the intersection of LLMs and the environmental domain, similar to other NLP subfields, efforts in this direction have been focused mostly on climate change (CC) topics rather than the broader environmental and ecological fields (which still include CC). Among the notable works, \cite{bulian2023assessing} and \cite{Zhu2023ClimateCF} propose evaluation frameworks for analyzing LLM responses to CC topics; \cite{koldunov2024local} built a prototype tool for localized climate-related data leveraging LLMs; \cite{leippold2024automated} developed an AI-based tool for fact-checking of CC claims utilizing an array of LLMs; \cite{Thulke2024ClimateGPTTA} introduced a family of domain-specific LLMs designed to synthesize interdisciplinary research on climate change. Unfortunately, these models require significant hardware resources to operate, making them difficult to access.

\section{Methodology}

\subsection{Baselines and Generative LMs employed} 

To establish baselines for assessing the added value of Large and Small Language Models (L/SLMs) in accurately classifying texts within the ecological domain, we reference the performance of six pre-trained BERT-based models. These models have been utilized in recent research across three classification tasks: binary Eco-Relevance classification, multilabel Environmental Impact Analysis, and Stance Detection\cite{grasso-etal-2024-ecoverse}. Key performance metrics from the studies include:
\begin{itemize}
    \item \textit{BERT and RoBERTa}\cite{devlin-etal-2019-bert,liu2020roberta}: Masked language models trained on large English corpora. In the Stance Detection task, RoBERTa achieved a high accuracy of 81.29\%, matching DistilRoBERTa, with BERT demonstrating high precision and F1 scores (95.09\% and 95.56\%, respectively).
    \item \textit{DistilRoBERTa} \cite{sanh2019distilbert}: An efficient adaptation of RoBERTa, showing the highest accuracy in the Eco-Relevance task at 89.43\%. It maintained competitive performance across all tasks, highlighting its efficiency.
    \item \textit{ClimateBERT}\cite{Webersinke2021ClimateBertAP}: Includes variants ClimateBertF, ClimateBertS, and ClimateBertS+D, pre-trained on climate-specific texts. In the Environmental Impact Analysis, ClimateBertS led in accuracy (78.62\%) and achieved the best F1 measure (54.67\%) among the ClimateBert variants.
\end{itemize}

These results set a comprehensive baseline for our study, allowing for a direct comparison of the effectiveness of L/SLMs in similar tasks\footnote{Here we report just the most informative performance score. The other scores are present in the referenced paper.}.
To account for a heterogeneous range of generative language models, we selected three different types of language models for our evaluation, drawing inspiration from \cite{Yu2023OpenCO}, to cover closed, open and small-scale models: 

\begin{itemize}
\item \textbf{GPT-3.5-Turbo-0125}. An advanced iteration within the GPT series developed by OpenAI \cite{achiam2023gpt}, representing high-capacity closed generative LLMs. Given its widespread use in both public and academic contexts, it serves as our benchmark for closed generative LLMs.
\item \textbf{Llama-2-13b-chat-hf}. A variant from Meta AI’s Llama-2 model series \cite{Touvron2023Llama2O}. Designed for conversational applications, it exemplifies mid-to-large open-domain language models with its capability for human-like text generation and comprehension.
\item \textbf{Gemma-2b-it}. A newly introduced small language model (SLM) by Google, derived from the Gemma family \cite{team2024gemma}. Gemma is a family of lightweight, state-of-the-art open models built from the same research and technology used to create the larger Gemini models.
\end{itemize}
\subsection{Classification Tasks and Prompt Design}
\begin{table*}[!ht]
\caption{Prompt for the EIA multilabel classification task, detailing instructions, labels, and a structured response format. Adapted versions were used for Eco-Relevance and Stance Detection tasks with task-specific modifications.}
    \centering
    \begin{tabularx}{\textwidth}{|X|} 
         \hline
         \textbf{Prompt} \\ 
         \hline
Perform a multilabel classification on the provided tweet from an ecological perspective. An instance could discuss topics that are either (potentially or explicitly) harmful or beneficial to the sustainability and well-being of the natural world.
Pay extreme attention to the mere content of the text and ignore completely the stance of who is writing the post.
Analyzing the content make inferences also on your background knowledge of green and environment topics. Consider paramount that this level is NOT a sentiment analysis.

The labels you must use for classification are:
"POS": The tweet content is potentially or explicitly beneficial to the sustainability and well-being of the natural world.
"NEU": The tweet content is not beneficial nor harmful to the sustainability and well-being of the natural world.
"NEG": The tweet content is potentially or explicitly harmful to the sustainability and well-being of the natural world.

Your response must strictly follow this format, pay attention to the order of the fields in the answer format independently from the assigned label, the MUST always be as follow:
LABEL: \textless{}[POS/NEU/NEG]\textgreater{}\; PROB\: POS\_PROB=\textless{}probability of Positive\textgreater{}, NEU\_PROB=\textless{}probability of Neutral\textgreater{}, NEG\_PROB=\textless{}probability of Negative\textgreater{}; EXP: \textless{}A natural language explanation detailing why the probabilities were assigned as such\textgreater{}.

Your answer must includes explicit probabilities for all the "POS", "NEU" and "NEG" labels, reflecting the likelihood of the tweet content being Beneficial (POS), Neutral (NEU) or Negative (NEG) towards the environment.
You must select ONE and ONLY ONE LABEL between POS, NEU and NEG for each single tweet.

For clarity, here are examples illustrating the expected format (Note: Examples provide the LABEL only for brevity):

- \textless{}tweet\_example\_1\textgreater{} [POS]
- \textless{}tweet\_example\_2\textgreater{} [POS]
- \textless{}tweet\_example\_3\textgreater{} [NEU]
- \textless{}tweet\_example\_4\textgreater{} [NEU]
- \textless{}tweet\_example\_5\textgreater{} [NEG]
- \textless{}tweet\_example\_6\textgreater{} [NEG]
Analyze and classify the following tweet according to these guidelines:
\textless{}tweet\textgreater{}\\ \hline
\end{tabularx}
\vspace{-7mm}
\label{tab:prompt}
\end{table*}
\subsubsection{Tasks Description} Numerous studies have showcased the effectiveness of few-shot learning in Language Models \cite{Brown2020LanguageMA,Clavi2023LargeLM}. In our study, we conducted a few-shot, three-layer text classification of tweets using the EcoVerse Dataset \cite{grasso-etal-2024-ecoverse}, comprised of 3k tweets, guiding the L/SLMs to perform classifications on three tasks.

\textit{Eco-Relevance}: The first task involves binary classification to identify texts related to Ecology. The labels for this level are \texttt{eco-related} or \texttt{not eco-related}.

\textit{Environmental Impact Analysis} (EIA): The second multilabel task determines, for eco-related tweets, whether the post conveys behaviors or events with 
\texttt{positive}, 
\texttt{negative}, or \texttt{neutral} impacts on the environment.

\textit{Stance Detection}: The third level identify the stance expressed by the tweet's author as \texttt{supportive}, \texttt{neutral}, or \texttt{skeptical/opposing} towards environmental causes.
\\For the few-shot examples, we randomly selected from the training set two examples for each label. Although more complex example sampling strategies have proven effective in scenarios where the demonstrative examples and the input text are significantly semantically divergent \cite{Sun2023TextCV}, we chose random sampling because the tweet topics in this case are relatively homogeneous. 

\subsubsection{Prompt Design}
Prompt Engineering is key to enhancing LLMs' precision, with prompt wording significantly influencing model's reasoning \cite{Clavi2023LargeLM}. By incorporating domain-specific knowledge, this technique helps LLMs match or exceed traditional models in efficiency, often with less data \cite{Deldjoo2023FairnessOC}. In text classification, the clarity of prompts is essential for generating categorical outputs. Our objective was to design prompts that yield discrete labels, allowing us to: (i) evaluate LLMs' classification performance on the EcoVerse dataset metrics like Accuracy and Precision; (ii) benchmark these results against BERT-based models for the same tasks; and (iii) assess the models' self-evaluation by examining the calibration of their verbalized confidence scores. Accordingly, the prompts were designed to produce direct outputs for quantitative analysis; be comparable with BERT-based models through a few-shot approach; and include verbalized confidence for calibration checks. To facilitate future qualitative analyses, especially regarding classification errors, prompts also requested a rationale for the model's choices. Table \ref{tab:prompt} illustrates the comprehensive prompt for the EIA multilabel task. Adaptations of this prompt for the Eco-Relevance and Stance Detection tasks were similarly structured but customized with task-specific wording and examples. 
\subsection{Experimental Setup}
We conducted our experiments on the Paperspace platform\footnote{\url{https://www.paperspace.com/}}, utilizing a configuration that includes an NVIDIA A100 GPU with 80GB of VRAM, 90GB of RAM, and a 12-core CPU. We employed the vLLM python interface\footnote{\url{https://blog.vllm.ai/2023/06/20/vllm.html}} to load the llama-2-13b-chat-hf and gemma-2b-it models. For the GPT-3.5-turbo model, we utilized OpenAI's APIs\footnote{\url{https://openai.com/blog/openai-api}}. We set a relatively low temperature of 0.3 for each model, capping the output at 512 tokens. This low temperature setting was chosen to reduce the models' "creativity," aiming for more consistent results on classification side. It's important to note that even setting the temperature to 0 can still result in some randomness in the models' outputs. To mitigate the impact of this unpredictability on performance and to obtain statistically significant results, similarly to what performed in other similar studies \cite{Motoki2023MoreHT} we executed each experiment 100 times for each of the three tasks and calculated the average metrics, which are presented in Table \ref{tab:combined_results}.




\section{Results and Discussion}
\subsection{Models' Classification Performance}
To evaluate the performance of the models, we utilized the following set of metrics: Accuracy, Precision, Recall, and the F1-score. For tasks involving multiple labels, we provided a comprehensive analysis by employing the macro-average versions of these metrics to offer a global perspective on model performance. The results for the three distinct tasks are presented and discussed below.
\subsubsection{Eco-Relevance Task}
As shown in Table \ref{tab:combined_results}, the GPT-3.5-Turbo-0125 model demonstrates superior performance compared to the other LLMs across all metrics, notably outperforming also BERT-based models in terms of recall. 
This suggests that GPT-3.5 is not only precise in identifying relevant instances but also excels at recognizing positive instances while minimizing false positives.
The Llama-2-13b-chat-hf model follows with commendable results, especially considering the recall measure where, along with GPT-3.5, surpass RoBERTa. 
This indicates its proficiency in identifying eco-relevant instances. However, its precision is notably lower, which is reflected in a marginally lower F1-score. This indicates that while Llama-2 is effective at detecting relevant instances, it is prone to incorrectly classifying non-eco-relevant instances as relevant.
Gemma-2b-it 
significantly lags behind the other two models across all metrics, possibly given its smaller scale. Despite the notable results of GPT-3.5 in this first classification task, 
it does not surpass the most performant language model of our selected baseline models, RoBERTa. 
\vspace{-9mm}
\setlength{\tabcolsep}{5pt} 
\begin{table}[h]
    \caption{Comparative results of models across the three different tasks. 
    }
    \centering    
    \begin{tabular}{| p{2cm} | l | c |c |c |c |}  
    \hline
    \rowcolor{lightgray}
    \textbf{Task} & \textbf{Model} & \textbf{Precision}& \textbf{Recall} & \textbf{F-1 score} & \textbf{Accuracy} \\
    \hline              
    \multirow{3}{*}{Eco-Relevance}
    &  $RoBERTa$ & \textbf{88.90\%} & 88.96\% & \textbf{88.93\%} & \textbf{88.87\%} \\
    & Llama-2 & 62,85\% & \textbf{95,37\%} & 75,76\% & 73,42\% \\
    & Gemma & 40,90\% & 38,49\% & 39,65\% & 48,98\% \\
    & GPT-3.5 & 72,12\% & \textbf{96,97\%} & 82,72\% & 82,34\% \\ \hline 

    \multirow{3}{*}{EIA}
    & $ClimBert_S$ & 51.81\% & 57.78\% & 54.67\% & 78.62\%\\
    & Llama-2 & 34,15\% & 30,73\% & 28,48\% & 31,98\% \\
    & Gemma & 40,42\% & 39,64\% & 39,69\% & 44,87\% \\
    & \textbf{GPT-3.5} & \textbf{68,56\%} & \textbf{66,69\%} & \textbf{66,75\%} & \textbf{84,07\%} \\ \hline

    \multirow{3}{*}{Stance Detect.}
    & \textbf{$BERT$} & \textbf{95.09\%} & \textbf{96.04\%} & \textbf{95.56\%} & \textbf{74.27\%} \\
    & Llama-2 & 26,33\% & 27,14\% & 26,47\% & 31,65\% \\
    & Gemma & 22,80\% & 37,30\% & 24,90\% & 27,60\% \\
    & GPT-3.5 & 56,85\% & 70,66\% & 56,28\% & 65,87\% \\
    \hline
    \end{tabular}  
    \label{tab:combined_results}
    
\end{table}
\vspace{-10mm}
\subsubsection{Environmental Impact Analyisis}
In this task, GPT-3.5 outperforms both the other generative models and the top performing model in EIA ClimateBert\_s, achieving the highest scores in all evaluated metrics. 
This notable performance could be attributed to its advanced reasoning capabilities which likely played a crucial role in investigating the complexities of EIA. This task can be challenging as it necessitates a comprehensive understanding that extends beyond linguistic patterns to include extra-linguistic contextual comprehension, meaning that a model must not only parse and understand the text but also relate it to broader environmental contexts.
\\\textbf{Stance Detection}
For the Stance Detection task the performances of all generative models appear to be less optimal, with GPT-3.5 leading among them but drastically underperforming compared to the BERT-like baseline models.
Llama-2 and Gemma performed significantly lower across all metrics, with Llama-2 achieving a slightly better balance between precision and recall. The underperformance of GPT-3.5 in this task might suggest that the specific characteristics and the fine-tuning of BERT models lend it an advantage in stance detection. 
It also indicates that the advanced reasoning capabilities and broader knowledge base of GPT-3.5, while beneficial for many tasks, may not always translate to superior performance in tasks where domain-specific training and optimization may play a crucial role.
\subsection{Calibration Evaluation}
For the calibration assessment, we employed the prompt-elicitation method, as detailed in Table 1, to extract verbalized confidence scores from the models, expressed as output tokens.
These scores are compared with actual precision measures to assess model calibration effectively \cite{Tian2023JustAF}. The evaluation process delves into how well the models' stated confidences align with their empirical precision across three tasks. This method provides insights into the consistency of models' self-assessed confidences over 100 iterations, highlighting any mismatches between their provided explanations and the probability scores. We categorized the scores into five probability bins: ['0-0.2', '0.2-0.4', '0.4-0.6', '0.6-0.8', '0.8-1.0'] then calibration was evaluated by comparing the mean probabilities with the observed precision for each bin. Below we detail the calibration performance of each model, focusing specifically on the task that demonstrated the most accurate calibration outcomes.

\textbf{GPT} demonstrates well-calibrated outputs across all tasks, with specific results for the stance detection task presented in Table \ref{tab:cal_gpt_stance}. The stance classification exhibits high confidence levels for the supportive and skeptical/opposing labels, albeit encountering some difficulty with the neutral stance. 
This results indicates that the probabilities are well-calibrated in each bin, indicating that GPT provides confident responses for correct answers, and similarly, manifests considerable uncertainty when the answers are incorrect.
\vspace{-5mm}
\begin{table}[!ht]
    \centering
    \caption{GPT-3.5 calibration results on Stance Detection Task. The table shows for each bin, their mean probability (Pr) for each label [support./neutral/skept-oppos.]. Alongside are presented the Precision (P) scores for each label.}
    \begin{tabular}{|c|c|c|c|c|}
    \hline
        \textbf{Bins} & \textbf{Bin Mean Pr } & \textbf{P\_sup} & \textbf{P\_neu} & \textbf{P\_ske} \\ \hline
        0.0 - 0.2 & [0,15/0,06/0,03] & 0 & 0 & 0 \\ \hline
        0.2 - 0.4 & [0,33/0,25/0,2] & 0 & 0 & 0 \\ \hline
        0.4 - 0.6 & [0,4/0,54/0,60] & 0 & 0 & 0 \\ \hline
        0.6 - 0.8 & [0,01/0,60/0,60] & 0 & 0,62 & 0,71 \\ \hline 
        0.8 - 1.0 & [0,91/0,01/0,82] & 0,83 & 0,34 & 0,88 \\ \hline
    \end{tabular}
    \label{tab:cal_gpt_stance}
\end{table}
\vspace{-5mm}

\textbf{LLaMA}'s calibration is notably moderate, with its most reliable performance observed in the stance detection task. Table \ref{tab:cal_llama_stance} reveals that model's probabilities are well-calibrated except for the \texttt{neutral} label 
where we observe poor calibration, most notably in the highest probability bin of 0.8 - 1.0 
indicating that the model confidently made incorrect decisions.\\
\\
\vspace{-8mm}
\begin{table}[!ht]
    \centering
    \caption{LLama-2 calibration results on Stance Detection Task}
    \begin{tabular}{|c|c|c|c|c|}
    \hline
        \textbf{Bins} & \textbf{Bin Mean Pr} & \textbf{P\_sup} & \textbf{P\_neu} & \textbf{P\_ske} \\ \hline
        0.0 - 0.2 & [0,09/0,09/0,06] & 0 & 0 & 0 \\ \hline
        0.2 - 0.4 & [0,21/0,25/0,2] & 0 & 0 & 0 \\ \hline
        0.4 - 0.6 & [0,47/0,4/0,0] & 0 & 0 & 0 \\ \hline
        0.6 - 0.8 & [0,6/0,61/0,80] & 0,55 & 0,28 & 0,85 \\ \hline
        0.8 - 1.0 & [0,8/0,8/0,8] & 0,88 & 0,57 & 0,89 \\ \hline
    \end{tabular}
    \vspace{-5mm}
    \label{tab:cal_llama_stance}
\end{table}


\textbf{Gemma} exhibits good calibration at the Eco-Relevance task, as shown in Table \ref{tab:cal_gemma_ecorel}. However, 
we were unable to compute it for subsequent levels as it failed to follow the prompt template, displaying too much inconsistency within the same iteration and across different iterations. 
This discrepancy compared to its larger counterparts can primarily be attributed to the architectural and training differences inherent to SLMs. 
To ensure a fair comparative analysis, we deliberately chose to maintain uniform testing conditions across all models. This approach was aimed at assessing each model’s ability to adapt to standardized tasks without necessitating model-specific prompt optimizations. The variations observed in Gemma's responses, particularly 
its inconsistent results in multi-label score distribution, underscore a critical insight: prompt engineering may need to be tailored to accommodate the 
limitations of smaller generative models. 
\vspace{-12mm}
\begin{table}[!ht]
    \centering
    \caption{Gemma calibration results on Eco-Relevance Task}
    \begin{tabular}{|c|c|c|c|}
    \hline
        \textbf{Bins} & \textbf{Bin Mean Pr} & \textbf{P\_eco\_rel} & \textbf{P\_not\_eco\_rel} \\ \hline
        0.0 - 0.2 & [0/0,03] & 0 & 0 \\ \hline
        0.2 - 0.4 & [0,26/0,25] & 0 & 0 \\ \hline
        0.4 - 0.6 & [0,49/0,50] & 0,59 & 0,54 \\ \hline
        0.6 - 0.8 & [0,74/0,74] & 0,9 & 0,88 \\ \hline
        0.8 - 1.0 & [0,92/0,94] & 0,96 & 0,95 \\ \hline
    \end{tabular}
    \vspace{-14mm}
    \label{tab:cal_gemma_ecorel}
\end{table}
\section{Conclusion}
In this study, we addressed three Research Questions (RQs) regarding the performance of advanced generative language models—GPT-3.5.Turbo-0125, Llama-2-13b-chat-hf, and Gemma-2b-it—in ecological domain tasks: Eco-Relevance, Environmental Impact Analysis (EIA), and Stance Detection. Our findings reveal:
\textbf{RQ1} showed variable effectiveness across models, with GPT demonstrating particular strength in Eco-Relevance, outperforming others.
\textbf{RQ2} assessed if these models could exceed a BERT-based benchmark. GPT excelled in the EIA task, yet no model consistently surpassed all baseline metrics. Gemma significantly underperformed compared to the other models.
\textbf{RQ3} evaluated the calibration of verbalized confidence scores, noting GPT's consistent reliability. Conversely, LLaMA and Gemma's calibration varied, indicating a need for refinement.
This study highlights the strengths and areas for improvement of generative models in environmental classification, contributing to the dialogue on NLP's role in addressing ecological and climate issues, and underscores the potential of LLMs in domain-specific tasks.

\bibliographystyle{splncs04}
\bibliography{bibliography}

\end{document}